\documentclass{article}

\usepackage{PRIMEarxiv}

\usepackage[utf8]{inputenc} 
\usepackage[T1]{fontenc}    
\usepackage{hyperref}       
\usepackage{url}            
\usepackage{booktabs}       
\usepackage{amsfonts}       
\usepackage{nicefrac}       
\usepackage{microtype}      
\usepackage{fancyhdr}       
\usepackage{graphicx}       
\graphicspath{{media/}{Figures/}}     

\usepackage{amsmath}
\usepackage{amssymb}
\usepackage{bm}
\usepackage{subcaption}
\usepackage{multirow}
\usepackage[normalem]{ulem}
\usepackage{enumitem}
\setlist{nosep, leftmargin=14pt}
\usepackage{tikz}

\title{Enhanced Diagnostic Performance via Large-Resolution Inference Optimization for Pathology Foundation Models

}

\author{
  Mengxuan Hu\thanks{Equal Contributions. $^{\dagger}$ Zihan Guan performed the work during the internship at Biometrics Research, Merck \& Co., Inc., Rahway, NJ, USA. $^{\ddagger}$ Corresponding Authors.}  \\
  University of Virginia \\
  Charlottesville, VA, USA\\
  \texttt{qtq7su@virginia.edu} \\
  \And
  Zihan Guan$^{\star, \dagger}$ \\
  University of Virginia \\
  Charlottesville, VA, USA\\
  \texttt{bxv6gs@virginia.edu} \\
  \And
  John Kang \\
  Merck \& Co., Inc. \\
  Biometrics Research \\
  Rahway, NJ, USA\\
  \texttt{jia.kang@merck.com} \\
  \AND
  Sheng Li$^{\ddagger}$ \\
  University of Virginia \\
  Charlottesville, VA, USA\\
  \texttt{shengli@virginia.edu} \\
  \And
  Zhongliang Zhou$^{\ddagger}$ \\
  Merck \& Co., Inc. \\
  Biometrics Research \\
  Rahway, NJ, USA\\
  \texttt{zhongliang.zhou@merck.com} \\
}

\begin{document}
\maketitle

\begin{abstract}
Despite their prominent performance on tasks such as ROI classification and segmentation, many pathology foundation models remain constrained by a specific input size e.g. $224 \times 224$, creating substantial inefficiencies when applied to whole-slide images (WSIs), which span thousands of resolutions. A na\"ive strategy is to either enlarge inputs or downsample the WSIs. However, enlarging inputs results in prohibitive GPU memory consumption, while downsampling alters the microns-per-pixel resolution and obscures critical morphological details. To overcome these limitations, we propose an space- and time- efficient inference strategy that sparsifies attention using spatially aware neighboring blocks and filters out non-informative tokens through global attention scores. This design substantially reduces GPU memory and runtime during high-resolution WSI inference while preserving and even improving the downstream performance, enabling inference at higher resolutions under the same GPU budget. The experimental results show that our method can achieves up to an 7.67\% improvement in the ROI classification and compatible results in segmentation.
\end{abstract}


\section{Introduction}

Tissue analysis is vital for cancer diagnosis and treatment, and traditional glass slides are increasingly being replaced by digital whole-slide images (WSIs). This shift enables computational pathology to move into routine clinical use~\cite{guanpharmadata, guan2023cohortgpt, liu2023pharmacygpt}. Recent advances are driven by foundation models~\cite{chen2024towards,xu2024whole,vorontsov2023virchow}—large deep neural networks trained on massive datasets with self-supervised learning. These models generate versatile data embeddings that generalize across diverse tasks including ROI-level classification, segmentation, and image retrieval, offering a clear advantage over diagnostic-specific methods limited by smaller, less varied pathology datasets.

\begin{figure}[!h]
    \centering
    \includegraphics[width=\linewidth]{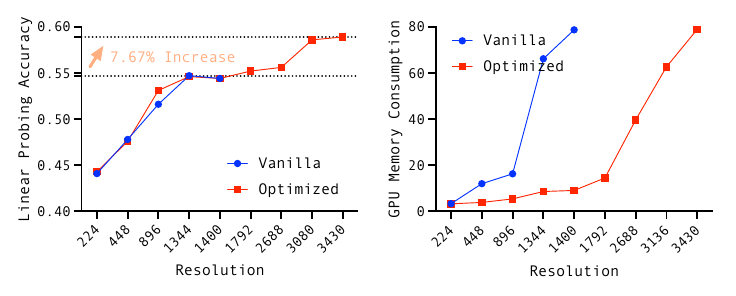}
    \caption{The left figure shows the interplay between the resolution and the linear probing accuracy on the PANDA dataset; The right figure shows the interplay between the resolution and the GPU memory consumption.}
    \label{fig:classification}
\end{figure}

Despite their strong performance, most pathology foundation models—commonly built on Vision Transformer (ViT) architectures with the DINO algorithm, typically trained for a specific input size $224 \times 224$ or $448 \times 448$. Whole-slide images (WSIs), however, contain information across thousands of resolutions, creating a substantial mismatch. A straightforward approach would be to either feed larger images directly into the model (since in principle Transformers can accept variable-length inputs) or to downsample them to match the trained input size. Both strategies inevitably introduce drawbacks. The former leads to quartic scaling of GPU memory usage and computation time with image resolution, while the later alters the microns-per-pixel (mpp) resolution, potentially obscuring critical morphological features such as cellular atypia~\cite{chen2024towards}. As shown in the right of Figure~\ref{fig:classification}, the GPU memory quadratically increases in the vanilla inference mode.

Our experimental results show that higher image resolutions generally lead to better downstream performance. Hence, we abandon the second solution and focus on improving the first one—feeding larger images directly into the model—by reducing GPU memory usage and runtime while maintaining or even improving performance. Specifically, we sparsify the attention matrix using spatially aware neighboring blocks, motivated by the observation that a cell's most relevant signals and interactions are typically concentrated within its immediate microenvironment. In addition, to further accelerate inference, we leverage global attention scores to rank image tokens and filter out those corresponding to histologically non-informative regions—such as empty background, or tissue preparation artifacts. This reduces computational overhead while simultaneously removing noisy signals that can hinder performance. Our empirical experiments indicate that the combined approach not only achieves a significant GPU memory reduction and efficiency improvement, but also achieves 7.67\% improvement in the classification task and compatible performance in the segmentation task.

To sum up, the contributions in this paper include:
\begin{itemize}
    \item \textbf{Resolution and Performance}: The first paper that explicitly discusses the interplay between input resolution and the downstream performance of pathology foundation models.
    \item \textbf{Inference Optimization}: The first paper that focuses on optimizing the inference of the pathology foundation models.
    \item \textbf{Promising Performance}: We achieve significant reduction in GPU memory and improvement in efficiency.
\end{itemize}

\section{Related Works}

\subsection{Digital Pathology Foundation Models}
Foundational vision models have significantly advanced digital pathology analysis. Notable examples include CTransPath~\cite{wang2022transformer}, which integrates convolutions within Transformers for spatial context; UNI~\cite{chen2024towards}, which uses hierarchical Transformers for multi-resolution processing; and GigaPath~\cite{xu2024whole}, which trains on gigapixel-scale images to capture fine-grained details. More recently, Virchow~\cite{vorontsov2023virchow} was introduced as a pathology-specific foundation model that excels at capturing clinically relevant features from diverse histopathology data. These models are predominantly trained with ViTs or similar architectures, whose computational cost scales quartically with image size. This severely limits their direct application to WSIs.

\subsection{Vision Transformers Inference Optimization} 
Pioneering studies on optimizing vision transformer inference for high-resolution images can be broadly categorized into two approaches. The first focuses on optimizing memory consumption within attention modules, as seen in methods like SparseAttention~\cite{zaheer2020big}, FlashAttention~\cite{dao2022flashattention}, and vLLM~\cite{kwon2023efficient}. The second approach aims to reduce computational overhead by pruning redundant attention calculations, exemplified by FastV~\cite{chen2024imageworth12tokens} and FastGen~\cite{ge2024model}. Although these optimizations have been shown to reduce model overhead without sacrificing performance in other settings, their design for histopathology images remains underexplored. To our knowledge, this is the first paper to specifically address inference optimization for pathology foundation models. Furthermore, our method is the first to simultaneously combine both memory optimization and computational pruning techniques, creating a more comprehensive solution than the previous work.

\section{Methodology}
\begin{figure}[!h]
    \centering
    \includegraphics[width=\linewidth]{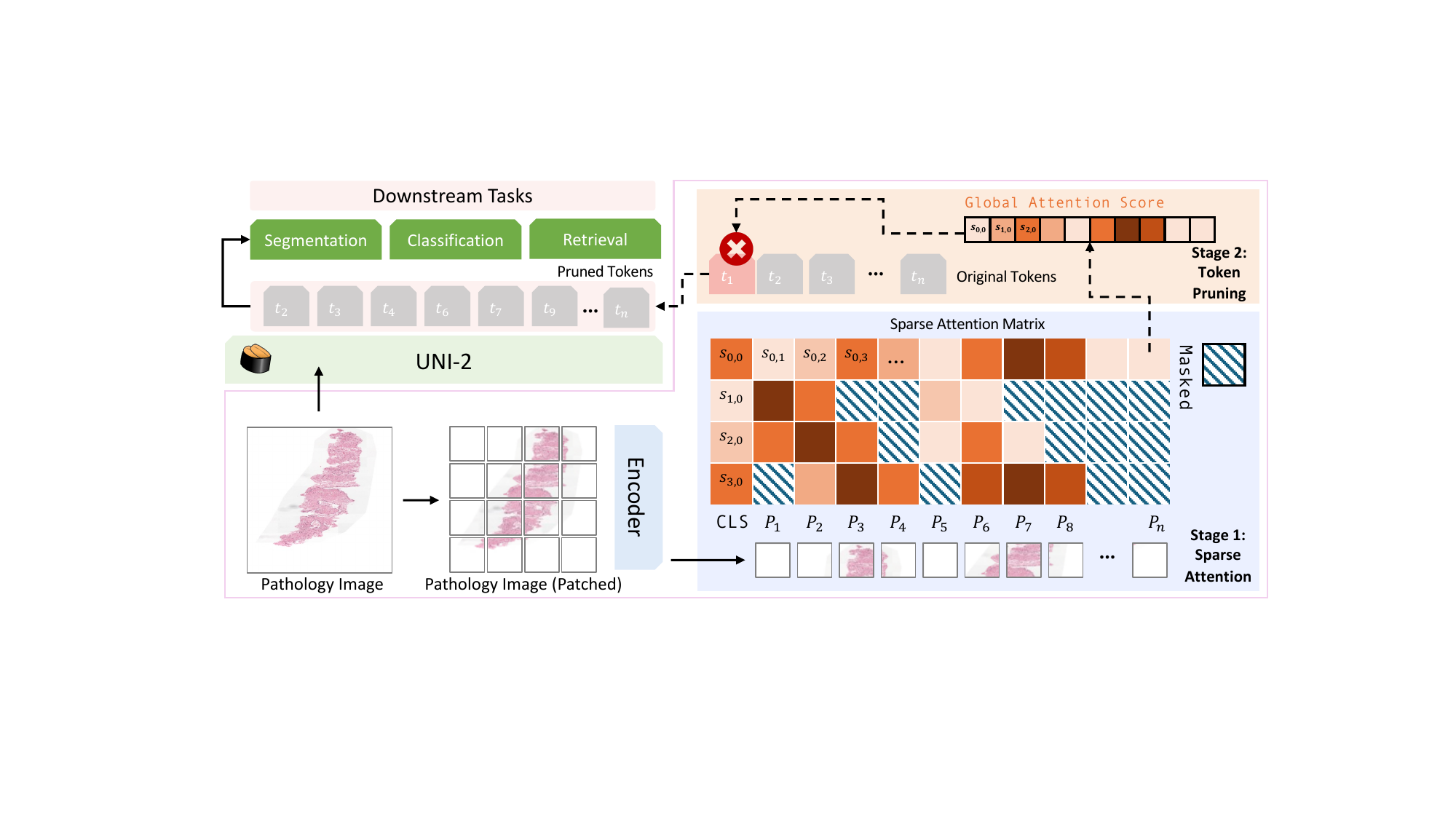}
    \caption{An overview of our inference optimization strategies. Sparse attention masks attention scores from distant tokens, while token pruning removes tokens with the lowest global attention scores.}
    \label{fig:uni_inference}
\end{figure}

We first introduce the intuitions for the inference optimizations in Section~\ref{sec:intuition}. Then, introduce the optimization strategies for both memory efficiency and time efficiency in Section~\ref{sec:memory_opt} and Section~\ref{sec:time_opt}. The overall optimization mechanism is depicted in Figure~\ref{fig:uni_inference}.

\subsection{The Trade-offs of Higher Resolutions}\label{sec:intuition}
Higher resolutions generally leads to better performance. Taken UNI-2 model as an example, when we increase the resolution of the input images, the performance on the downstream classification task progressively increases (shown in the Figure~\ref{fig:classification} Left). However, the benefits come at a significant cost of memory consumption: The GPU memory consumption increases quartically along with the resolution (Right). Given the great potential benefits of using larger resolutions as inputs and the bottlenecks posed by the GPU memory consumption, it is natural to conjecture that, \textit{can we push the Pareto frontier of the performance and the GPU memory consumption by optimizing the model inference process?}

\subsection{Memory-Efficiency Optimization}\label{sec:memory_opt}
The primary computational bottleneck is the quadratic memory complexity ($\mathcal{O}(N^2)$) of the self-attention mechanism, where the attention matrix multiplications can consume significant amount of GPU memories. This is further exemplified in ViT, as N itself scales quadratically with input resolution. To alleviate the memory burden, our approach is motivated by the observation that in histopathology images, a patch's most informative context lies within its immediate microenvironment. The inherent heterogeneity of tissue organization makes local neighborhoods particularly relevant for interpretation, whereas distant regions often contribute little. Conventional attention mechanisms, however, still allocate substantial computation to these less informative areas.

Therefore, we employ a sparse attention mechanism, inspired by BigBird~\cite{zaheer2020big}. Instead of computing the full attention matrix $A \in \mathbb{R}^{N\times N}$, where $N$ is the number of tokens, we sparsify the interactions using a hybrid approach:
\begin{itemize}
    \item \textbf{Local Windowed Attention}: Each token attends only to tokens corresponding to its spatially adjacent patches in the original pixel space. This efficiently captures high-fidelity local features and context.
    \item \textbf{Global Attention}: A few designated "global" tokens (e.g., CLS) attend to all other tokens in the sequence. This ensures that critical long-range dependencies are maintained across the entire image.
\end{itemize}

This combined strategy allows the model to build a rich understanding of the input by focusing computational resources on meaningful local interactions while still preserving essential global context.

\subsection{Time-Efficiency Optimization}\label{sec:time_opt}

Our second optimization targets time efficiency. Even with sparse attention, the sequence length N can be very large for high-resolution images, leading to slow processing speeds. To accelerate computation, we introduce a dynamic token pruning stage, motivated by FastViT~\cite{chen2024imageworth12tokens}. This method identifies and removes computationally redundant tokens—such as those corresponding to empty background regions or tissue artifacts—based on their contribution. Specifically, we derive an importance score for each of the N tokens from the global attention weights, forming a score vector $a \in \mathbb{R}^{N}$. We then rank all tokens by their importance and permanently discard a proportion $p$ (where $0 \leq p < 1$) with the lowest scores. To ensure the scores contain meaningful semantic information before pruning, we apply the pruning after several initial layers have processed the data (e.g., after the fourth layer). This strategy dynamically shortens the sequence length for all subsequent layers, substantially reducing overall computation time. Furthermore, by pruning uninformative tokens, the model focuses more effectively on semantically relevant regions, leading to improved downstream performance.

\section{Experimental Results}

\subsection{Experimental Setups}

\paragraph{Tasks and Datasets} We evaluate our method on two typical tasks for pathology foundation models: PANDA~\cite{bulten2022artificial} dataset for cancer pattern classification and SegPath~\cite{komura2023restaining} for cancer histology segmentation. Detailed descriptions of the task setups are provided in the following sections.

\paragraph{Settings} For all the experiments, the default window size is 8, and the pruning ratio is 0.6 unless otherwise specified. We adopt UNI-2~\cite{chen2024towards} as the backbone foundation model. However, the methodology can be easily integrated with other foundation models~\cite{xu2024whole, chen2024towards, lu2024visual}. We manually resize the input images to simulate different image resolutions (e.g., 224, 448). All the experiments are conducted on an A100 (80GB) GPU.

\subsection{Cancer Classification Task}

\paragraph{Setup.} The PANDA dataset is used for the task of classifying cancer tissue samples into one of five International Society of Urological Pathology (ISUP) scores. Following~\cite{chen2024towards}, we adopt two evaluation strategies when using pathology foundation models for this task: linear probing and K-Nearest Neighbor (KNN). Linear probing involves training a simple linear classifier on top of features extracted from the frozen, pre-trained foundation model. In contrast, KNN assigns a label to a test sample based on the majority vote of its 20 nearest neighbors in the feature space, using Euclidean distance. For both strategies, we report accuracy, the weighted F1-score, and the Area Under the Receiver Operating Characteristic curve (AUROC).

\paragraph{Main Results.} 
Table~\ref{tab:main_classification} gives a full comparison of the performance, time efficiency, and GPU memory consumption between the vanilla inference workflow and our optimized method. As shown, \uline{when input with the same fixed resolution, our method significantly reduces time and GPU memory consumption while maintaining or even achieving better performance}. The benefits in performance gain can be partially explained by our design of the inference strategy: the spatial-aware sparse attention largely maintains the necessary information for decision-making, and the pruned uninformative tokens guide the model to better capture the semantic information of the input images. Moreover, it is observed that \uline{on the standard A100 GPU of 80 GB GPU memory, our method eventually can accommodate inputs with 3430 resolutions, while the vanilla inference can accommodate inputs with only 1400 resolutions. This increase in input resolutions finally led to a significant increase in performance and efficiency in image processing for the real-world workflow.}

\begin{table}[!t]
    \centering
    \small
    \begin{tabular}{cc|ccccc}
    \toprule
        \multirow{3}*{Resolution} & \multirow{3}*{Method} & \multicolumn{3}{c}{Linear Probing} & \multicolumn{2}{c}{Time} \\
          \cmidrule(lr){3-5} \cmidrule(lr){6-7}
         & & Accuracy & Weighted F1 & AUROC & Time & Memory \\
         \midrule
         \multirow{2}*{224} &  Vanilla & 0.441 & 0.433 & \textbf{0.751} & \textcolor{red}{2,241} & \textcolor{red}{3.35}\\
          &  Ours &  \textbf{0.443} & \textbf{0.436} & 0.750 & \textcolor{green!60}{2,181} & \textcolor{green!60}{3.32}\\
          \midrule
          \multirow{2}*{448} &  Vanilla & \textbf{0.478} & \textbf{0.474} & \textbf{0.777} & \textcolor{green!60}{2,248} & \textcolor{red}{12.03}\\
          &  Ours & 0.476 & 0.469 & 0.776 & \textcolor{red}{2,285} & \textcolor{green!60}{3.90}\\
          \midrule
          \multirow{2}*{896} &  Vanilla & 0.516 & 0.510 & 0.798 & \textcolor{red}{6,583} & \textcolor{red}{16.32}\\
          &  Ours & \textbf{0.531} & \textbf{0.522} & \textbf{0.816} & \textcolor{green!60}{3,476} &  \textcolor{green!60}{5.41} \\
          \midrule
          \multirow{2}*{1344} &  Vanilla & \textbf{0.547} & \textbf{0.540} & 0.828 & \textcolor{red}{21,703} & \textcolor{red}{66.18}\\
          &  Ours &  0.546 & 0.538 & \textbf{0.832} & \textcolor{green!60}{7,668} & \textcolor{green!60}{8.66}\\
          \midrule
          \multirow{2}*{1400} &  Vanilla & \textbf{0.544} & \textbf{0.536} & 0.822 & \textcolor{red}{24,818} & \textcolor{red}{78.73} \\
          &  Ours & \textbf{0.544} & 0.534 & \textbf{0.830} & \textcolor{green!60}{8,360} & \textcolor{green!60}{9.12} \\
          \midrule
          \multirow{2}*{1792} &  Vanilla & - & - & - & - &  -\\
          &  Ours & 0.552 & 0.545 & 0.841 & \textcolor{green!60}{13,588} & \textcolor{green!60}{14.50} \\
          \midrule
          \multirow{2}*{2688} &  Vanilla & - & - & - & - & - \\
          & Ours & 0.556 & 0.545 & 0.843 & \textcolor{green!60}{30,315} & \textcolor{green!60}{39.61} \\
          \midrule
          \multirow{2}*{3136} &  Vanilla & - & - & - & - & -\\
          &  Ours &  0.586 & 0.576 & 0.849 & \textcolor{green!60}{41,368} & \textcolor{green!60}{62.63}\\
          \midrule
          \multirow{2}*{3430} &  Vanilla & - & - & - & - & -\\
          &  Ours & 0.589 & 0.574 & 0.852 & \textcolor{green!60}{89,640} &\textcolor{green!60}{78.87}\\
    \bottomrule
    \end{tabular}
    \caption{Performance comparison between our method and the vanilla UNI-2 model on the classification task. \textit{Shorter} runtime and \textit{lower} GPU usage are highlighted in \textcolor{green!60}{green}. \textit{Longer} runtime and \textit{higher} GPU usage are highlighted in \textcolor{red}{red}.}
    \label{tab:main_classification}
\end{table}

\paragraph{Ablation Studies.} 
Table~\ref{tab:pruning_ratio} and Table~\ref{tab:window_size} present ablation studies on the pruning ratio $p$ and window size $c$, respectively, when the image size is of resolution 3134 (i.e., 224 $\times$ 14). The results indicate that performance remains consistently stable across different choices of $p$ and $c$.

\begin{table}[!t]
    \centering
    \small
    \begin{tabular}{c|cccccc}
    \toprule
         & \multicolumn{3}{c}{Linear Probing} & \multicolumn{3}{c}{KNN} \\
          \cmidrule(lr){2-4} \cmidrule(lr){5-7}
         & Accuracy & F1 & AUROC & Accuracy & Bal. Acc. & F1 \\
         \midrule
        $p = 0.4$ & 0.576 & 0.569 & 0.852 & 0.479 & 0.415 & 0.463 \\
        $p = 0.5$ & 0.581 & 0.572 & 0.849 & 0.476 & 0.408 & 0.457\\
        $p = 0.6$ & 0.586 & 0.576 & 0.849 & 0.480 & 0.411 & 0.461\\
        $p = 0.7$ & 0.578 & 0.568 & 0.850 & 0.486 & 0.416 & 0.467\\
        $p = 0.8$ & 0.572 & 0.562 & 0.849 & 0.473 & 0.402 & 0.452\\
    \bottomrule
    \end{tabular}
    \caption{Ablation studies on pruning ratio $p$}
    \label{tab:pruning_ratio}
\end{table}

\begin{table}[!t]
    \centering
    \small
    \begin{tabular}{c|cccccc}
    \toprule
         & \multicolumn{3}{c}{Linear Probing} & \multicolumn{3}{c}{KNN} \\
          \cmidrule(lr){2-4} \cmidrule(lr){5-7}
         & Accuracy & F1 & AUROC & Accuracy & Bal. Acc. & F1 \\
         \midrule
         $w = 2$ & 0.585 & 0.577 & 0.849 & 0.465 & 0.396 & 0.444 \\
        $w = 8$ & 0.586 & 0.576 & 0.849 & 0.480 & 0.411 & 0.461\\
        $w = 16$ & 0.579 & 0.569 & 0.849 & 0.477 & 0.408 & 0.458 \\
        $w = 64$ & 0.573 & 0.561 & 0.849 & 0.463 & 0.400 & 0.443 \\
    \bottomrule
    \end{tabular}
    \caption{Ablation studies on windows size $w$}
    \label{tab:window_size}
\end{table}

\begin{figure}[t!]
    \centering
    \begin{subfigure}[b]{0.48\textwidth}
        \centering
        \includegraphics[width=\linewidth]{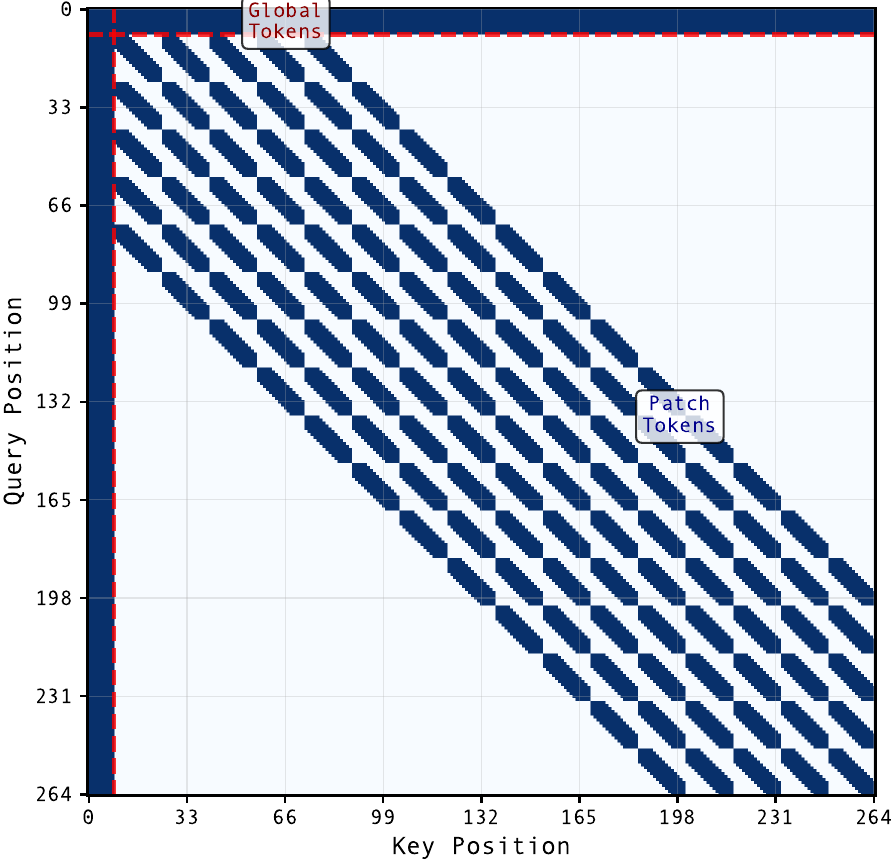}
        \caption{Sparsed Attention Matrix. The light blue color denotes the masked attention values.}
        \label{fig:sparse_attention}
    \end{subfigure}%
    \hfill
    \begin{subfigure}[b]{0.48\textwidth}
        \centering
        \includegraphics[width=\linewidth]{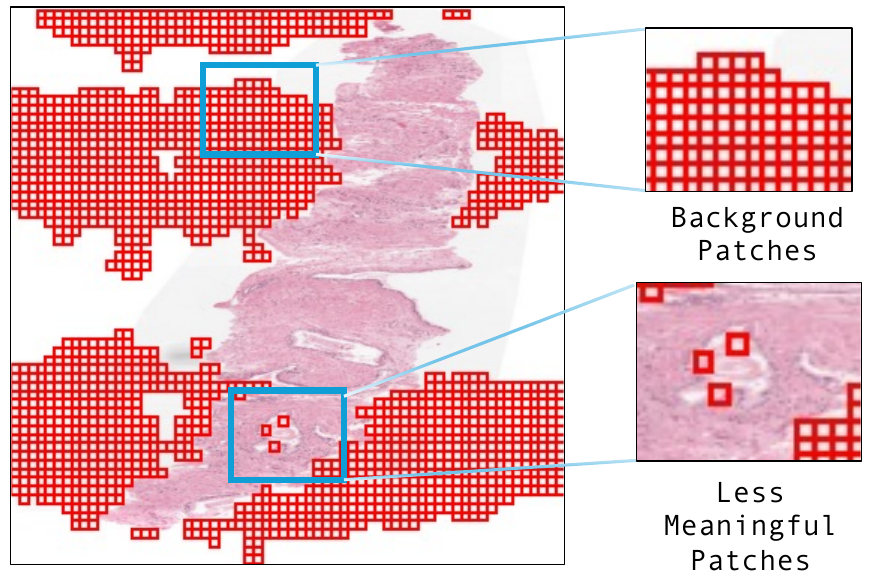}
        \caption{Pruned image tokens mapped to the pixel-space. The red rectangles \textcolor{red}{\rule{1em}{1ex}} indicates pruned image patches (each measuring $14 \times 14$ pixels).}
        \label{fig:image_prune}
    \end{subfigure}
    \caption{Qualitative examples of the sparse attention matrix and pruned image tokens.}
\end{figure}

\paragraph{Qualitative Examples on Image Token Pruning.} Figure~\ref{fig:sparse_attention} visualizes the sparsified attention matrix. Figure~\ref{fig:image_prune} shows the pruned image tokens mapped back to the original pixel space, where the red rectangle \textcolor{red}{\rule{1em}{1ex}} indicates a pruned image token. The pruning mainly results from empty background patches or less informative regions.

\subsection{Segmentation Task} 

\paragraph{Setup.} SegPath is used for the tasks of segmenting eight major cell types in tumor tissue: epithelial cells, smooth muscle cells, red blood cells, endothelial cells, leukocytes, lymphocytes, plasma cells, and myeloid cells. Each image is of size $984 \times 984$. We employ the Mask2Former model~\cite{cheng2022masked}, fine-tuned with a Vision Transformer Adapter (ViT-Adapter) to segment eight major cell types within tumor tissues using the SegPath dataset. The training hyperparameters follow the setups in~\cite{yan2025pathorchestra}. Following the prior work~\cite{chen2024towards, yan2025pathorchestra, guan2024badsam}, we report metrics such as Mean Pixel Accuracy (MPA), class pixel accuracy (CPA), Intersection over Union (IoU), and Dice, using the averaged performance across the eight cell types in the SegPath. It is noted that Mask2Former requires feature maps that preserve the original image's spatial structure; we cannot apply the time-efficiency optimization described in Section~\ref{sec:time_opt} to the features extracted by the UNI-2 model. Therefore, we do not report the time consumption in the Table~\ref{tab:main_segmentation}.

\paragraph{Main Results} The results for the segmentation task show a similar pattern as in the classification task (see Table~\ref{tab:main_segmentation}). However, since segmentation relies on global spatial information,  sparsifying the attention matrix inevitably drops semantical information that is potentially meaningful to the segmentation. Therefore, the performance gap between our method and the vanilla method is relatively larger.

\begin{table}[!t]
    \centering
    \small
    \begin{tabular}{cc|cccc}
    \toprule
        \multirow{2}*{Resolution} & \multirow{2}*{Method} & \multicolumn{3}{c}{Segmentation Performance} & \multirow{2}*{Memory} \\
          \cmidrule(lr){3-5}
         & & MPA & IoU & Dice &  \\
         \midrule
         \multirow{2}*{112} & Vanilla & 0.632 & 0.553 & 0.657 & \textcolor{red}{7.15} \\
          &  Ours & 0.642 & 0.458 & 0.587 & \textcolor{green!60}{6.93} \\
          \midrule
          \multirow{2}*{224} & Vanilla & 0.727 & 0.553 & 0.678 & \textcolor{red}{12.99} \\
          &  Ours & 0.661 & 0.467 & 0.596 & \textcolor{green!60}{12.77}\\
          \midrule
          \multirow{2}*{336} & Vanilla  & 0.742 & 0.517 & 0.650 & \textcolor{red}{24.09}\\
          &  Ours & 0.678 & 0.497 & 0.623 & \textcolor{green!60}{22.79}\\
          \midrule
          \multirow{2}*{448} & Vanilla & 0.749 & 0.579 & 0.702 & \textcolor{red}{42.33}\\
          &  Ours & 0.709 & 0.565 & 0.686 & \textcolor{green!60}{35.38} \\
          \midrule
          \multirow{2}*{560} & Vanilla & 0.743 & 0.568 & 0.701 & \textcolor{red}{60.03}\\
          &  Ours & 0.709 & 0.558 & 0.678 & \textcolor{green!60}{52.55}  \\
    \bottomrule
    \end{tabular}
    \caption{Performance comparison between our method and the vanilla UNI-2 model on the segmentation task. \textit{Lower} GPU usage is highlighted in \textcolor{green!60}{green}. \textit{Higher} GPU usage is highlighted in \textcolor{red}{red}.}
    \label{tab:main_segmentation}
\end{table}

\paragraph{Ablation Studies} Table~\ref{tab:window_size_segmentation} presents the ablation studies on the window size $c$, when the image size is of resolution $560$. The results indicate that the performance tends to improve when the window size is larger, which aligns with the intuition that the segmentation task relies more on the global features.

\begin{table}[!t]
    \centering
    \small
    \begin{tabular}{c|ccc}
    \toprule
         & MPA & IoU & Dice \\
         \midrule
         $w = 2$ & 0.686 & 0.547 & 0.66\\
         $w = 8$ & 0.709 & 0.558 & 0.678\\
         $w = 16$ & 0.710 & 0.559 & 0.680 \\
         $w = 64$ & 0.723 & 0.576 & 0.696 \\
    \bottomrule
    \end{tabular}
    \caption{Ablation studies on windows size $w$.}
    \label{tab:window_size_segmentation}
\end{table}

\begin{figure}[!t]
    \centering
    \includegraphics[width=0.8\linewidth]{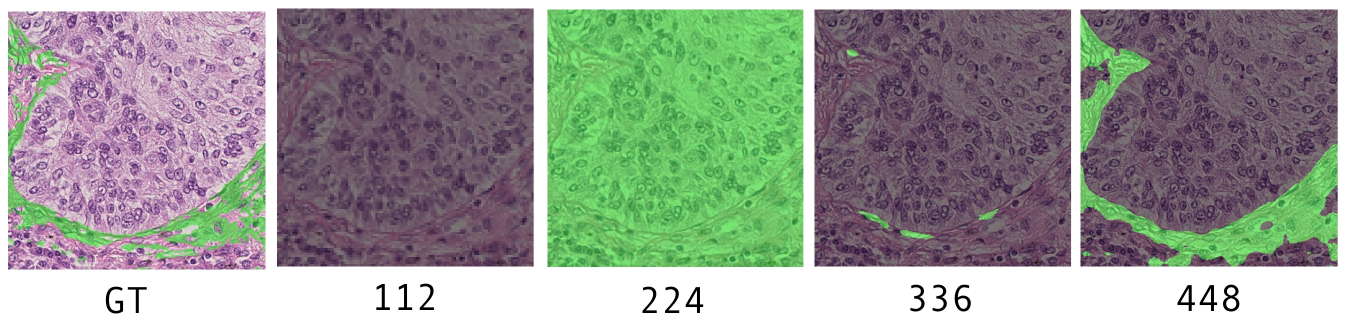}
    \caption{Qualitative examples of the UNI-2's performance on the segmentation task when the input pixel varies.}
    \label{fig:segmentation_qualitative}
\end{figure}

\paragraph{Qualitative Examples on Segmentation Performance} Figure~\ref{fig:segmentation_qualitative} presents the qualitative examples of the UNI-2's performance on the segmentation task when the input pixel varies from $112 \times 112$ to $448 \times 448$. As shown, the predicted segmentation mask aligns more precisely with the ground-truth segmentation mask when the resolution increases.

\section{Conclusion and Future Directions}
In this paper, we propose an inference optimization method for digital pathology foundation models. The results on ROI classification and segmentation tasks demonstrate the effectiveness of our method. Despite the promising performance, there are still many directions to explore, e.g., What other methods could be specially designed for the vision language model (VLM)-based pathology foundation models?

\section*{Acknowledgments}
This project is supported by Merck BARDS Academic Collaboration Grant.

\bibliographystyle{unsrt}  
\bibliography{references}

@article{xu2024whole,
  title={A whole-slide foundation model for digital pathology from real-world data},
  author={Xu, Hanwen and Usuyama, Naoto and Bagga, Jaspreet and Zhang, Sheng and Rao, Rajesh and Naumann, Tristan and Wong, Cliff and Gero, Zelalem and Gonz{\'a}lez, Javier and Gu, Yu and others},
  journal={Nature},
  volume={630},
  number={8015},
  pages={181--188},
  year={2024},
  publisher={Nature Publishing Group UK London}
}

@misc{chen2024imageworth12tokens,
      title={An Image is Worth 1/2 Tokens After Layer 2: Plug-and-Play Inference Acceleration for Large Vision-Language Models}, 
      author={Liang Chen and Haozhe Zhao and Tianyu Liu and Shuai Bai and Junyang Lin and Chang Zhou and Baobao Chang},
      year={2024},
      eprint={2403.06764},
      archivePrefix={arXiv},
      primaryClass={cs.CV},
      url={https://arxiv.org/abs/2403.06764}, 
}

@article{bulten2022artificial,
  title={Artificial intelligence for diagnosis and Gleason grading of prostate cancer: the PANDA challenge},
  author={Bulten, Wouter and Kartasalo, Kimmo and Chen, Po-Hsuan Cameron and Str{\"o}m, Peter and Pinckaers, Hans and Nagpal, Kunal and Cai, Yuannan and Steiner, David F and Van Boven, Hester and Vink, Robert and others},
  journal={Nature medicine},
  volume={28},
  number={1},
  pages={154--163},
  year={2022},
  publisher={Nature Publishing Group US New York}
}

@article{zaheer2020big,
  title={Big bird: Transformers for longer sequences},
  author={Zaheer, Manzil and Guruganesh, Guru and Dubey, Kumar Avinava and Ainslie, Joshua and Alberti, Chris and Ontanon, Santiago and Pham, Philip and Ravula, Anirudh and Wang, Qifan and Yang, Li and others},
  journal={Advances in neural information processing systems},
  volume={33},
  pages={17283--17297},
  year={2020}
}

@article{komura2023restaining,
  title={Restaining-based annotation for cancer histology segmentation to overcome annotation-related limitations among pathologists},
  author={Komura, Daisuke and Onoyama, Takumi and Shinbo, Koki and Odaka, Hiroto and Hayakawa, Minako and Ochi, Mieko and Herdiantoputri, Ranny Rahaningrum and Endo, Haruya and Katoh, Hiroto and Ikeda, Tohru and others},
  journal={Patterns},
  volume={4},
  number={2},
  year={2023},
  publisher={Elsevier}
}

@inproceedings{cheng2022masked,
  title={Masked-attention mask transformer for universal image segmentation},
  author={Cheng, Bowen and Misra, Ishan and Schwing, Alexander G and Kirillov, Alexander and Girdhar, Rohit},
  booktitle={Proceedings of the IEEE/CVF conference on computer vision and pattern recognition},
  pages={1290--1299},
  year={2022}
}

@article{chen2024towards,
  title={Towards a general-purpose foundation model for computational pathology},
  author={Chen, Richard J and Ding, Tong and Lu, Ming Y and Williamson, Drew FK and Jaume, Guillaume and Song, Andrew H and Chen, Bowen and Zhang, Andrew and Shao, Daniel and Shaban, Muhammad and others},
  journal={Nature Medicine},
  volume={30},
  number={3},
  pages={850--862},
  year={2024},
  publisher={Nature Publishing Group US New York}
}

@article{lu2024visual,
  title={A visual-language foundation model for computational pathology},
  author={Lu, Ming Y and Chen, Bowen and Williamson, Drew FK and Chen, Richard J and Liang, Ivy and Ding, Tong and Jaume, Guillaume and Odintsov, Igor and Le, Long Phi and Gerber, Georg and others},
  journal={Nature Medicine},
  volume={30},
  number={3},
  pages={863--874},
  year={2024},
  publisher={Nature Publishing Group US New York}
}

@article{yan2025pathorchestra,
  title={Pathorchestra: A comprehensive foundation model for computational pathology with over 100 diverse clinical-grade tasks},
  author={Yan, Fang and Wu, Jianfeng and Li, Jiawen and Wang, Wei and Lu, Jiaxuan and Chen, Wen and Gao, Zizhao and Li, Jianan and Yan, Hong and Ma, Jiabo and others},
  journal={arXiv preprint arXiv:2503.24345},
  year={2025}
}

@article{vorontsov2023virchow,
  title={Virchow: A million-slide digital pathology foundation model},
  author={Vorontsov, Eugene and Bozkurt, Alican and Casson, Adam and Shaikovski, George and Zelechowski, Michal and Liu, Siqi and Severson, Kristen and Zimmermann, Eric and Hall, James and Tenenholtz, Neil and others},
  journal={arXiv preprint arXiv:2309.07778},
  year={2023}
}

@article{wang2022transformer,
  title={Transformer-based unsupervised contrastive learning for histopathological image classification},
  author={Wang, Xiyue and Yang, Sen and Zhang, Jun and Wang, Minghui and Zhang, Jing and Yang, Wei and Huang, Junzhou and Han, Xiao},
  journal={Medical image analysis},
  volume={81},
  pages={102559},
  year={2022},
  publisher={Elsevier}
}

@inproceedings{kwon2023efficient,
  title={Efficient Memory Management for Large Language Model Serving with PagedAttention},
  author={Woosuk Kwon and Zhuohan Li and Siyuan Zhuang and Ying Sheng and Lianmin Zheng and Cody Hao Yu and Joseph E. Gonzalez and Hao Zhang and Ion Stoica},
  booktitle={Proceedings of the ACM SIGOPS 29th Symposium on Operating Systems Principles},
  year={2023}
}

@article{dao2022flashattention,
  title={Flashattention: Fast and memory-efficient exact attention with io-awareness},
  author={Dao, Tri and Fu, Dan and Ermon, Stefano and Rudra, Atri and R{\'e}, Christopher},
  journal={Advances in neural information processing systems},
  volume={35},
  pages={16344--16359},
  year={2022}
}

@inproceedings{
ge2024model,
title={Model Tells You What to Discard: Adaptive {KV} Cache Compression for {LLM}s},
author={Suyu Ge and Yunan Zhang and Liyuan Liu and Minjia Zhang and Jiawei Han and Jianfeng Gao},
booktitle={The Twelfth International Conference on Learning Representations},
year={2024},
url={https://openreview.net/forum?id=uNrFpDPMyo}
}

@inproceedings{guanpharmadata,
  title={PharmaData-Agent: A Specialized Agent for Pharmaceutical Data Analysis},
  author={Guan, Zihan and Wang, Hanyin and Zhou, Zhongliang and Zhou, Qiaohui and Tao, Peining and Ma, Junshui},
  booktitle={The Second Workshop on GenAI for Health: Potential, Trust, and Policy Compliance},
  year={2025},
}

@article{guan2023cohortgpt,
  title={Cohortgpt: An enhanced gpt for participant recruitment in clinical study},
  author={Guan, Zihan and Wu, Zihao and Liu, Zhengliang and Wu, Dufan and Ren, Hui and Li, Quanzheng and Li, Xiang and Liu, Ninghao},
  journal={arXiv preprint arXiv:2307.11346},
  year={2023}
}

@article{liu2023pharmacygpt,
  title={Pharmacygpt: The ai pharmacist},
  author={Liu, Zhengliang and Wu, Zihao and Hu, Mengxuan and Zhao, Bokai and Zhao, Lin and Zhang, Tianyi and Dai, Haixing and Chen, Xianyan and Shen, Ye and Li, Sheng and others},
  journal={arXiv preprint arXiv:2307.10432},
  year={2023}
}

@inproceedings{guan2024badsam,
  title={Badsam: Exploring security vulnerabilities of sam via backdoor attacks (student abstract)},
  author={Guan, Zihan and Hu, Mengxuan and Zhou, Zhongliang and Zhang, Jielu and Li, Sheng and Liu, Ninghao},
  booktitle={Proceedings of the AAAI Conference on Artificial Intelligence},
  volume={38},
  number={21},
  pages={23506--23507},
  year={2024}
}

\end{document}